\documentclass[11pt]{article}

\usepackage[utf8]{inputenc}
\usepackage[T1]{fontenc}
\usepackage{amsmath,amssymb}
\usepackage{booktabs}
\usepackage{multirow}
\usepackage{graphicx}
\usepackage{hyperref}
\usepackage[margin=1in]{geometry}
\usepackage{natbib}
\usepackage{xcolor}
\usepackage{array}
\usepackage{caption}

\hypersetup{
    colorlinks=true,
    linkcolor=blue,
    citecolor=blue,
    urlcolor=blue
}

\title{When Context Hurts: The Crossover Effect of Knowledge Transfer on Multi-Agent Design Exploration}

\author{Saranyan Vigraham \\
Meta}

\date{}

\begin{document}

\maketitle

\begin{abstract}
The prevailing assumption in agent orchestration is that more context is better. We test this on multi-agent software design across 10 tasks, 7 context-injection conditions, and over 2,700 runs, and find a \emph{crossover effect}: the same artifact type improves design exploration on some tasks (up to 20$\times$ tradeoff coverage) and actively degrades it on others (up to 46\% reduction). On several tasks, an \emph{irrelevant} document performs as well as or better than every relevant artifact. The direction is predicted by a single measurable variable---\emph{baseline exploration} without context---with Pearson $r = -0.82$ ($p < 0.001$). Probing the mechanism by manipulating convergence pressure through prompt design reveals two distinct regimes: convergence driven by training data priors (\emph{natural}) responds to artifact disruption, while convergence driven by explicit instructions (\emph{induced}) does not. The implication is that context injection should be conditional, not universal: one no-context trial is a cheap diagnostic that predicts whether knowledge artifacts will help or hurt a given task.
\end{abstract}


\section{Introduction}

The AI-assisted software engineering community has an untested assumption about context: more is better. Every major advance in the field has increased the amount of information available to AI agents. Context windows have grown from 4K to 1M+ tokens. Retrieval-augmented generation injects relevant documents before every query. Agent frameworks feed prior conversation, code, documentation, and design artifacts into each interaction. The assumption underlying all of these advances is that relevant context improves output quality.

For code generation---producing correct implementations of specified behavior---this assumption is well-supported. An agent given a function signature, type definitions, and usage examples writes better code than one working from a bare prompt \citep{chen2021evaluating, li2022competition}.

But software engineering is not just code generation. It is design: evaluating alternatives, weighing tradeoffs, and making architectural decisions that shape the solution space. For design work, the relationship between context and quality is more complex---and, as we show, sometimes inverted.

This paper reports a controlled experiment in which we measure how knowledge artifacts from one agent team affect the design exploration of a different team solving the same problem. We test seven types of context injection---from raw deliberation transcripts to polished design documents to code---across 10 software design tasks. The tasks range from textbook problems (rate limiter, LRU cache) to domain-specific challenges (Kubernetes operator, database storage engine).

\subsection{The Crossover Effect}

Our central finding is a \emph{crossover effect}: the same knowledge artifact type produces opposite effects depending on the task. Anti-pattern documents increase tradeoff coverage by 20$\times$ on rate limiter (from 0.033 to 0.700, $p < 0.001$) but decrease it by 9\% on Kubernetes operator (from 0.475 to 0.431). Deliberation transcripts increase coverage by 17$\times$ on rate limiter (from 0.033 to 0.592, $p < 0.001$) but decrease it by 46\% on Kubernetes operator (from 0.475 to 0.256, $p < 0.001$).

The crossover is not random. It is predicted by a single variable: \emph{baseline exploration}---the tradeoff coverage that agents achieve with no injected context. When baseline is near zero, agents converge on a default solution and context disrupts this convergence. When baseline is moderate to high, agents are already exploring and context anchors them to a specific framing.

This finding has practical implications for the design of agent orchestration systems. Context injection should be conditional, gated on a measurement of baseline exploration. Unconditional context injection---the current industry default---will sometimes improve and sometimes degrade the quality of agent design work.

\subsection{Contributions}

\begin{enumerate}
    \item \textbf{The crossover effect in knowledge transfer.} We provide the first empirical demonstration that the same knowledge artifact type can improve or degrade multi-agent design exploration, with the direction predictable from a measurable task property. This challenges the assumption that context is uniformly beneficial.

    \item \textbf{Baseline exploration as a diagnostic metric.} We propose measuring design tradeoff coverage without context injection as a diagnostic for whether context will help or hurt. This metric is cheap to compute (run one baseline trial) and strongly predictive of artifact effectiveness ($r = -0.82$, $p < 0.001$).

    \item \textbf{Natural vs induced convergence.} We show that convergence pressure can be manipulated through prompt design, but that artifacts which disrupt natural convergence (driven by training data priors) fail to disrupt induced convergence (driven by explicit instructions). This mechanistic distinction reveals that the crossover effect depends on \emph{why} agents converge, not just \emph{that} they converge.

    \item \textbf{A reproducible evaluation method for design exploration.} We introduce \emph{direct evaluation}, a rubric-anchored method for measuring design tradeoff coverage that discriminates between conditions where standard evaluation (code quality, test passage) cannot.
\end{enumerate}

\subsection{Reproducibility}

All experiments use publicly available tools and models. Agent teams are 5 instances of Claude Sonnet 4 with distinct personas, run via the Claude Code CLI. Each condition uses 20 independent trials. All results reported are from runs that completed successfully ($>$30 seconds runtime), with error counts noted. No results are cherry-picked across tasks---all 10 tasks and all 7 conditions are reported regardless of whether they support our hypothesis.

\section{Related Work}

\subsection{Context and Performance in LLMs}

The relationship between context length and performance has been studied extensively. \citet{liu2024lost} find that LLMs struggle to use information in the middle of long contexts (``lost in the middle''), suggesting that more context is not straightforwardly better. \citet{shi2023large} show that irrelevant context can degrade reasoning performance. Our work extends these findings from single-model question answering to multi-agent design, and identifies a regime (convergent tasks) where context \emph{improves} performance---creating a crossover rather than a uniform degradation.

\subsection{Anchoring in LLMs}

Anchoring bias---disproportionate weighting of initial or salient information---has been documented in LLMs \citep{jones2022anchoring, talboy2024challenging}. Our crossover effect is consistent with an anchoring interpretation: injected context anchors agents to specific solutions. When agents would otherwise explore freely, this narrows the search space (harmful). When agents would otherwise converge on a single default, the anchor pulls them toward an alternative (helpful---disruption). This dual role of anchoring is novel in the LLM literature.

\subsection{Multi-Agent Software Engineering}

ChatDev \citep{qian2024chatdev}, MetaGPT \citep{hong2024metagpt}, and SWE-Debate \citep{yang2025swedebate} evaluate multi-agent systems by output quality---code correctness, test passage, issue resolution. None measure design space exploration. We argue that exploration quality is an orthogonal evaluation dimension that reveals dynamics invisible to output-focused metrics: a team can produce correct code while exploring only a fraction of the design space.

\subsection{Knowledge Transfer and Design Rationale}

Design rationale capture \citep{dutoit2006rationale} records why decisions were made alongside the decisions themselves. The assumption has always been that captured rationale helps future teams. Our data complicates this: captured rationale (design documents, tradeoff lists) helps only when the receiving team would not have explored those alternatives independently. When the team is already exploring, rationale acts as an anchor rather than an enabler.

\section{Method}

\subsection{Experimental Design}

The experiment has three phases:

\textbf{Phase 1---Baseline survey (10 tasks $\times$ 20 trials = 200 runs):} Agent teams solve each design task with no injected context. Establishes baseline exploration per task.

\textbf{Phase 2---Transfer experiment (10 tasks $\times$ 7 conditions $\times$ 20 trials = 1,400 runs):} A \emph{seed team} solves the task and produces deliberation artifacts. A \emph{transfer team} receives one artifact type and solves the same task. We measure whether the artifact changes tradeoff coverage.

\textbf{Phase 3---Convergence manipulation (2 tasks $\times$ 4 pressure levels $\times$ 7 conditions $\times$ 20 trials = 1,120 runs):} We vary convergence pressure via prompt design across four levels (none, mild, strong, extreme) and test whether artifact effects change as predicted.

Total: over 2,700 runs, each producing a full multi-agent deliberation and evaluated independently.

\subsection{Tasks}

Ten software design tasks across two categories:

\textbf{General SWE (5 tasks):} Well-known CS problems with extensive training data coverage.

\begin{table}[h]
\centering
\small
\begin{tabular}{lcp{8cm}}
\toprule
Task & Tradeoffs & Example decisions \\
\midrule
Rate limiter & 6 & Algorithm choice, build vs reuse, deployment model, lock granularity, cleanup, storage abstraction \\
LRU cache & 5 & Data structure, thread safety, write policy, eviction callbacks, size vs count limits \\
Task queue & 6 & Priority implementation, worker pool, retry strategy, persistence, error handling, concurrency model \\
Pub/sub broker & 8 & Delivery guarantees, persistence, ordering, consumer groups, backpressure, serialization, push vs pull, partitioning \\
Distributed scheduler & 10 & Push/pull, centralized/distributed, DAG handling, execution guarantees, priority, checkpointing, scheduling approach, resource awareness, leader election, batch vs streaming \\
\bottomrule
\end{tabular}
\end{table}

\textbf{Domain-specific (5 tasks):} Require specialized knowledge beyond general CS.

\begin{table}[h]
\centering
\small
\begin{tabular}{lcp{8cm}}
\toprule
Task & Tradeoffs & Example decisions \\
\midrule
Kubernetes operator & 8 & Framework choice, reconciliation strategy, leader election, CRD schema, failover detection, state management, update strategy, finalizer design \\
Database storage engine & 8 & LSM vs B-tree, WAL design, compaction, bloom filters, compression, memory management, index granularity, concurrency control \\
ML training pipeline & 8 & Data/model parallelism, gradient sync, checkpointing, mixed precision, data loading, framework choice, communication backend, batch scaling \\
Video streaming & 8 & ABR algorithm, segment duration, transcoding strategy, CDN routing, live vs VOD architecture, codec selection, DRM, manifest format \\
Network congestion control & 8 & Loss vs delay-based, RTT estimation, slow start, recovery strategy, ACK processing, fairness model, multiplexing, pacing \\
\bottomrule
\end{tabular}
\end{table}

Each task's known tradeoff list was validated by running seed trials and checking that the evaluator correctly identifies tradeoffs agents discuss using these labels (Section~\ref{sec:evaluation}).

\subsection{Agent Configuration}

\begin{itemize}
    \item \textbf{Model:} Claude Sonnet 4 (claude-sonnet-4-20250514)
    \item \textbf{Team size:} 5 agents with distinct personas (assigned via personality profiles derived from code review history)
    \item \textbf{Orchestration:} SA (Speed + Autonomy) mode---agents work in parallel, no inter-agent communication during work phase, followed by synthesis
    \item \textbf{Temperature:} 0.5
    \item \textbf{Trials:} 20 per condition, each trial fully independent (new agent instantiation, no shared state)
\end{itemize}

\subsection{Artifact Conditions}

Seven conditions represent different ways of transferring knowledge from a seed team to a transfer team:

\begin{table}[h]
\centering
\small
\begin{tabular}{clp{9cm}}
\toprule
ID & Condition & Description \\
\midrule
C1 & Transcript & Full, unedited deliberation record from all 5 seed agents \\
C2 & Topology & Extracted tradeoff list: named decisions with options and reasoning \\
C3 & Design doc & LLM-generated polished design document summarizing the seed team's approach \\
C4 & Anti-patterns & Rejected alternatives only---what the seed team considered and decided against \\
C5 & Code & Final implementation files from the seed team, no reasoning \\
C6 & Baseline & No artifact injected (control) \\
C7 & Irrelevant & An unrelated technical document (priming control) \\
\bottomrule
\end{tabular}
\end{table}

Artifacts are injected into the transfer team's task prompt as an appendix: \emph{``A previous team worked on this problem. Here is [artifact type] from their work: [content].''}

\subsection{Convergence Pressure Manipulation (Phase 3)}

We manipulate convergence by varying the specificity of the task prompt across four levels. The same task is presented with increasing prescription:

\begin{table}[h]
\centering
\small
\begin{tabular}{llp{6cm}}
\toprule
Level & Manipulation & Example (distributed scheduler) \\
\midrule
None & Open-ended & ``Design a distributed task scheduler'' \\
Mild & Name the standard approach & ``Most implementations use a centralized coordinator with worker polling'' \\
Strong & Prescribe the approach & ``Use a centralized coordinator with worker polling. Follow this pattern.'' \\
Extreme & Provide code skeleton & ``Here is a canonical implementation: [code]. Extend it.'' \\
\bottomrule
\end{tabular}
\end{table}

We apply this to two tasks: LRU cache (natural baseline = 0.540, exploratory) and distributed scheduler (natural baseline = 0.310, moderate). If convergence is the mechanism, increasing pressure should lower baseline exploration and potentially change the artifact effect direction.

\subsection{Evaluation: Direct Tradeoff Assessment}
\label{sec:evaluation}

Standard evaluation metrics---code correctness, test passage rates, implementation completeness---do not discriminate between convergent and exploratory behavior. A team that builds a working rate limiter using only token bucket scores identically to a team that deliberated over five algorithms.

We use \emph{direct evaluation}: an evaluator LLM (Claude Sonnet 4) reads each team's full deliberation record and, for each known tradeoff, answers:

\begin{enumerate}
    \item \textbf{Was this tradeoff discussed?} (yes/no)
    \item \textbf{What is the evidence?} (quote or paraphrase from the deliberation)
\end{enumerate}

Coverage ratio = (tradeoffs discussed) / (known tradeoffs for this task).

The evaluator also identifies \emph{novel tradeoffs}---design tensions discussed by the team that are not in the known list. This ensures we are not penalizing teams that explore in dimensions we did not anticipate.

\textbf{Validation:} We validated the evaluation method by checking that: (a) tradeoff names match agent vocabulary (recalibrated for rate\_limiter after initial mismatch); (b) the evaluator produces evidence quotes, not just labels; (c) known-positive cases (seed runs with rich deliberation) score above 0.5.

\section{Results: The Crossover Effect}

\subsection{Baseline Exploration Varies Dramatically Across Tasks}

With no injected context (C6---baseline), tradeoff coverage ranges from 0.033 to 0.540:

\begin{table}[h]
\centering
\caption{Baseline tradeoff coverage (no context injected), sorted ascending. $n = 20$ per task.}
\label{tab:baselines}
\begin{tabular}{lccc}
\toprule
Task & Baseline & $n$ & Tradeoffs \\
\midrule
Rate limiter & 0.033 & 20 & 6 \\
Pub/sub broker & 0.281 & 20 & 8 \\
Task queue & 0.308 & 20 & 6 \\
Distributed scheduler & 0.310 & 20 & 10 \\
ML training pipeline & 0.356 & 20 & 8 \\
Network congestion control & 0.400 & 20 & 8 \\
Database storage engine & 0.406 & 20 & 8 \\
Video streaming & 0.406 & 20 & 8 \\
Kubernetes operator & 0.475 & 20 & 8 \\
LRU cache & 0.540 & 20 & 5 \\
\bottomrule
\end{tabular}
\end{table}

Rate limiter teams discuss virtually no tradeoffs---they build ``the standard token bucket'' without considering sliding windows, fixed windows, or distributed alternatives. LRU cache teams, working on an equally familiar problem, discuss over half the known tradeoffs unprompted.

The difference is not complexity or familiarity. Both are well-known problems. The difference is whether a \emph{dominant default solution} exists in the model's training data. Rate limiters have one canonical answer (token bucket). LRU caches do not---the choice between OrderedDict and doubly-linked-list-plus-hashmap is genuinely open, as are thread safety and write policy decisions.

\subsection{The Full Effect Matrix}

The complete $10 \times 7$ matrix shows the delta from baseline for every task-condition pair. Positive values indicate the artifact improved exploration; negative values indicate it reduced exploration. Statistically significant results ($p < 0.05$, Welch's $t$-test) are marked with asterisks.

\begin{table}[h]
\centering
\caption{Full effect matrix (delta from baseline). $^*p < 0.05$, $^{**}p < 0.01$, $^{***}p < 0.001$. Welch's $t$-test, $n = 20$ per cell.}
\label{tab:effect_matrix}
\small
\begin{tabular}{lcccccc}
\toprule
Task (baseline) & Transcript & Topology & Design Doc & Anti-pat. & Code & Irrelevant \\
\midrule
Rate limiter (0.033) & $+0.558^{***}$ & $+0.175^*$ & $+0.158^*$ & $\mathbf{+0.667^{***}}$ & $+0.333^{***}$ & $+0.358^{***}$ \\
Pub/sub (0.281) & $-0.050$ & $+0.013$ & $-0.081$ & $+0.025$ & $-0.013$ & $+0.006$ \\
Task queue (0.308) & $-0.075$ & $+0.150^*$ & $-0.092$ & $+0.000$ & $-0.100^*$ & $+0.017$ \\
Dist.\ sched.\ (0.310) & $-0.130^*$ & $-0.005$ & $-0.070$ & $-0.035$ & $-0.080$ & $-0.045$ \\
ML pipeline (0.356) & $-0.125^{**}$ & $+0.075$ & $-0.062$ & $-0.050$ & $-0.100^*$ & $+0.087$ \\
Net.\ CC (0.400) & $-0.106$ & $-0.062$ & $-0.044$ & $+0.019$ & $-0.144^{**}$ & $+0.025$ \\
DB engine (0.406) & $-0.138^{**}$ & $+0.050$ & $-0.100^*$ & $+0.044$ & $-0.013$ & $-0.019$ \\
Video (0.406) & $-0.131^*$ & $-0.050$ & $-0.044$ & $+0.037$ & $-0.056$ & $+0.137$ \\
K8s operator (0.475) & $-0.219^{***}$ & $-0.119$ & $-0.150^*$ & $-0.044$ & $-0.150^*$ & $-0.006$ \\
LRU cache (0.540) & $-0.060$ & $-0.080$ & $-0.010$ & $+0.030$ & $-0.050$ & $-0.020$ \\
\bottomrule
\end{tabular}
\end{table}

The pattern is visible at a glance: the top row (rate limiter, baseline = 0.033) is entirely positive---every artifact helps. The bottom rows (baseline $> 0.4$) are predominantly negative---most artifacts hurt. Transcript and code show the widest swings between the two regimes, while anti-patterns and irrelevant context show the mildest negative effects on exploratory tasks.

\subsection{Transcripts: Highest Upside, Highest Downside}

Deliberation transcripts produce the most extreme effects in both directions:

\begin{table}[h]
\centering
\caption{Transcript artifact effect by task, sorted by delta. $n = 20$ per cell.}
\label{tab:transcripts}
\begin{tabular}{lccccr}
\toprule
Task & Baseline & Transcript & Delta & $p$ & $d$ \\
\midrule
Rate limiter & 0.033 & 0.592 & $\mathbf{+0.558}$ & $< 0.001$ & $+2.71$ \\
Pub/sub broker & 0.281 & 0.231 & $-0.050$ & $0.436$ & $-0.25$ \\
LRU cache & 0.540 & 0.480 & $-0.060$ & $0.242$ & $-0.37$ \\
Task queue & 0.308 & 0.233 & $-0.075$ & $0.173$ & $-0.43$ \\
Net.\ CC & 0.400 & 0.294 & $-0.106$ & $0.084$ & $-0.55$ \\
ML pipeline & 0.356 & 0.231 & $\mathbf{-0.125}$ & $0.005$ & $-0.88$ \\
Dist.\ scheduler & 0.310 & 0.180 & $\mathbf{-0.130}$ & $0.014$ & $-0.77$ \\
Video streaming & 0.406 & 0.275 & $\mathbf{-0.131}$ & $0.044$ & $-0.64$ \\
DB engine & 0.406 & 0.269 & $\mathbf{-0.138}$ & $0.006$ & $-0.87$ \\
K8s operator & 0.475 & 0.256 & $\mathbf{-0.219}$ & $< 0.001$ & $-1.14$ \\
\bottomrule
\end{tabular}
\end{table}

Transcripts significantly improve exploration on rate limiter ($+0.558$, $d = 2.71$) but significantly degrade it on 5 of the remaining 9 tasks, with Kubernetes operator showing the worst effect ($-0.219$, $d = -1.14$). The raw, unresolved nature of transcripts amplifies both directions: their ambiguity disrupts convergence (good when agents are stuck) but introduces noise and anchors agents to specific debate framings (bad when agents are already exploring productively). Transcripts are the most dangerous artifact type---highest upside and highest downside.

\subsection{Anti-Patterns: Strongest Disruptor, Mildest Harm}

\begin{table}[h]
\centering
\caption{Anti-pattern artifact effect by task, sorted by delta. $n = 20$ per cell.}
\label{tab:antipatterns}
\begin{tabular}{lccccr}
\toprule
Task & Baseline & Anti-patterns & Delta & $p$ & $d$ \\
\midrule
Rate limiter & 0.033 & 0.700 & $\mathbf{+0.667}$ & $< 0.001$ & $+3.41$ \\
DB engine & 0.406 & 0.450 & $+0.044$ & $0.441$ & $+0.24$ \\
Video streaming & 0.406 & 0.444 & $+0.037$ & $0.552$ & $+0.19$ \\
LRU cache & 0.540 & 0.570 & $+0.030$ & $0.500$ & $+0.21$ \\
Pub/sub broker & 0.281 & 0.306 & $+0.025$ & $0.689$ & $+0.13$ \\
Net.\ CC & 0.400 & 0.419 & $+0.019$ & $0.743$ & $+0.10$ \\
Task queue & 0.308 & 0.308 & $+0.000$ & $1.000$ & $+0.00$ \\
Dist.\ scheduler & 0.310 & 0.275 & $-0.035$ & $0.515$ & $-0.21$ \\
K8s operator & 0.475 & 0.431 & $-0.044$ & $0.504$ & $-0.21$ \\
ML pipeline & 0.356 & 0.306 & $-0.050$ & $0.312$ & $-0.32$ \\
\bottomrule
\end{tabular}
\end{table}

Anti-patterns produce the largest single improvement (rate limiter: $+0.667$, $d = 3.41$, $p < 0.001$) but the smallest degradation on exploratory tasks. No individual negative effect reaches significance. Anti-patterns are the safest artifact type: they name what \emph{not} to do, which disrupts a default solution but provides weaker anchoring than a specific approach or implementation.

\subsection{Code: Strongest Anchor}

\begin{table}[h]
\centering
\caption{Code artifact effect by task, sorted by delta. $n = 20$ per cell.}
\label{tab:code}
\begin{tabular}{lccccr}
\toprule
Task & Baseline & Code & Delta & $p$ & $d$ \\
\midrule
Rate limiter & 0.033 & 0.367 & $\mathbf{+0.333}$ & $< 0.001$ & $+1.11$ \\
DB engine & 0.406 & 0.394 & $-0.013$ & $0.816$ & $-0.07$ \\
Pub/sub broker & 0.281 & 0.269 & $-0.013$ & $0.849$ & $-0.06$ \\
LRU cache & 0.540 & 0.490 & $-0.050$ & $0.424$ & $-0.25$ \\
Video streaming & 0.406 & 0.350 & $-0.056$ & $0.450$ & $-0.24$ \\
Dist.\ scheduler & 0.310 & 0.230 & $-0.080$ & $0.100$ & $-0.52$ \\
ML pipeline & 0.356 & 0.256 & $\mathbf{-0.100}$ & $0.030$ & $-0.69$ \\
Task queue & 0.308 & 0.208 & $\mathbf{-0.100}$ & $0.034$ & $-0.67$ \\
Net.\ CC & 0.400 & 0.256 & $\mathbf{-0.144}$ & $0.008$ & $-0.83$ \\
K8s operator & 0.475 & 0.325 & $\mathbf{-0.150}$ & $0.016$ & $-0.76$ \\
\bottomrule
\end{tabular}
\end{table}

Code helps only on rate limiter ($+0.333$, $p < 0.001$) and significantly degrades exploration on 4 tasks. Code is the strongest anchor because it presents a concrete, complete solution. Agents adopt the implementation rather than deliberate over alternatives. On exploratory tasks, this narrows the design space more than any other artifact type.

\subsection{Irrelevant Context Outperforms Relevant Context}

Perhaps the most striking result: on several tasks, an \emph{irrelevant} document outperforms every \emph{relevant} artifact.

\textbf{ML Training Pipeline:}

\begin{table}[h]
\centering
\caption{ML training pipeline---irrelevant context outperforms all relevant artifacts. $n = 20$ per cell.}
\label{tab:irrelevant}
\begin{tabular}{lccc}
\toprule
Condition & Coverage & vs Baseline & $p$ \\
\midrule
Irrelevant context & 0.444 & $\mathbf{+0.087}$ & 0.062 \\
Topology & 0.431 & $+0.075$ & 0.129 \\
Baseline & 0.356 & --- & --- \\
Anti-patterns & 0.306 & $-0.050$ & 0.312 \\
Design doc & 0.294 & $-0.062$ & 0.205 \\
Code & 0.256 & $-0.100$ & 0.030 \\
Transcript & 0.231 & $\mathbf{-0.125}$ & 0.005 \\
\bottomrule
\end{tabular}
\end{table}

An unrelated document increases coverage by 0.087 while every relevant artifact except topology \emph{decreases} coverage, with transcript significantly so ($p = 0.005$). Relevant content performs \emph{worse} than irrelevant content because it anchors agents to a specific framing. Irrelevant content provides generic disruption without content-specific anchoring.

This pattern repeats across the benchmark. On video streaming, irrelevant context ($+0.137$) is the only artifact that meaningfully outperforms baseline. On distributed scheduler, irrelevant context ($-0.045$) is the least harmful artifact while transcript ($-0.130$, $p = 0.014$) and code ($-0.080$) perform substantially worse. The consistency of this pattern---irrelevant beats relevant on exploratory tasks---suggests that the \emph{content} of context matters less than whether agents were already exploring.

\subsection{The Anchoring Mechanism}

The crossover effect is consistent with an \emph{anchoring-as-dual-force} interpretation:

\begin{enumerate}
    \item \textbf{When agents converge} (low baseline): The model has a strong prior for the ``standard'' solution. The injected artifact acts as a \emph{counter-anchor}---it introduces an alternative framing that competes with the default, forcing deliberation. Anti-patterns are the strongest counter-anchor because they explicitly name alternatives (``don't do X'' implies X exists as an option). Transcripts work similarly because their unresolved tensions resist convergence.

    \item \textbf{When agents explore} (high baseline): The model does not have a strong prior, so it naturally considers multiple approaches. The injected artifact \emph{becomes} the anchor---agents latch onto the specific framing, approach, or vocabulary in the artifact, reducing the diversity of their exploration. Code is the strongest anchor because it presents a concrete, complete solution. Transcripts are a moderate anchor because they describe specific approaches even while debating them.

    \item \textbf{Irrelevant context as weak disruption:} Irrelevant documents do not provide content-specific anchoring. They may partially disrupt the default prior (by occupying context window space or reducing model confidence) without introducing a competing anchor. This explains why irrelevant context performs between baseline and relevant artifacts on convergent tasks (weaker disruption) but sometimes outperforms relevant artifacts on exploratory tasks (no harmful anchoring).
\end{enumerate}

\subsection{Baseline Exploration Predicts Artifact Direction}

We compute the \emph{best artifact effect} for each task (maximum delta from baseline across all 7 conditions) and plot against baseline:

\begin{table}[h]
\centering
\caption{Best artifact effect vs baseline, sorted by baseline. $n = 20$ per cell.}
\label{tab:baseline_prediction}
\begin{tabular}{ccll}
\toprule
Baseline & Best Delta & Best Condition & Task \\
\midrule
0.033 & $\mathbf{+0.667}$ & Anti-patterns & Rate limiter \\
0.281 & $+0.025$ & Anti-patterns & Pub/sub broker \\
0.308 & $+0.150$ & Topology & Task queue \\
0.310 & $-0.005$ & Topology & Dist.\ scheduler \\
0.356 & $+0.087$ & Irrel.\ context & ML training \\
0.400 & $+0.025$ & Irrel.\ context & Network CC \\
0.406 & $+0.137$ & Irrel.\ context & Video streaming \\
0.406 & $+0.050$ & Topology & DB storage engine \\
0.475 & $-0.006$ & Irrel.\ context & K8s operator \\
0.540 & $+0.030$ & Anti-patterns & LRU cache \\
\bottomrule
\end{tabular}
\end{table}

The relationship is strongly correlated: lower baseline $\rightarrow$ larger artifact effect (Pearson $r = -0.821$, $p < 0.001$; Spearman $\rho = -0.624$, $p = 0.024$). Below baseline $\approx 0.1$, artifacts produce massive improvements. Above $\approx 0.3$, the best artifact produces marginal gains. Above $\approx 0.5$, no artifact helps. Notably, for tasks with baseline $> 0.3$, the best artifact is often irrelevant context---suggesting that even the best-case ``help'' on exploratory tasks is generic disruption, not content-specific knowledge transfer.

\subsection{Statistical Significance}

We test each of the 60 condition-task pairs (10 tasks $\times$ 6 non-baseline conditions) using Welch's $t$-test (unequal variances) and report Cohen's $d$ for effect size. All tests are two-tailed with $n = 20$ per cell.

\textbf{Significant improvements ($p < 0.05$):} 7 of 60 pairs, concentrated on rate limiter. Anti-patterns ($d = 3.41$, $p < 0.001$), transcript ($d = 2.71$, $p < 0.001$), irrelevant context ($d = 2.22$, $p < 0.001$), code ($d = 1.11$, $p < 0.001$), topology ($d = 0.80$, $p = 0.011$), design doc ($d = 0.77$, $p = 0.015$). The only non-rate-limiter improvement: topology on task queue ($d = 0.75$, $p = 0.017$). Every artifact type, including irrelevant context, significantly improves exploration on the most convergent task.

\textbf{Significant degradations ($p < 0.05$):} 11 of 60 pairs, spread across 6 tasks. Transcript is the most reliably harmful: it significantly degrades Kubernetes operator ($d = -1.14$, $p < 0.001$), ML training pipeline ($d = -0.88$, $p = 0.005$), database storage engine ($d = -0.87$, $p = 0.006$), distributed scheduler ($d = -0.77$, $p = 0.014$), and video streaming ($d = -0.64$, $p = 0.044$). Code significantly degrades network congestion control ($d = -0.83$, $p = 0.008$), Kubernetes operator ($d = -0.76$, $p = 0.016$), ML training pipeline ($d = -0.69$, $p = 0.030$), and task queue ($d = -0.67$, $p = 0.034$).

\textbf{Baseline-direction correlation:} Baseline exploration vs average artifact delta: Pearson $r = -0.843$, $p < 0.001$; Spearman $\rho = -0.624$, $p = 0.024$. Transcript-specific: $r = -0.833$, $p < 0.001$.

The crossover effect is not an artifact of noisy data. On rate limiter, every artifact type produces a large, highly significant improvement. On exploratory tasks, transcripts and code produce significant degradation with medium-to-large effect sizes. The baseline-direction correlation holds across both parametric and non-parametric tests.

\section{Results: Convergence Manipulation}

The Phase 2 results establish a strong correlation between baseline exploration and artifact direction. Phase 3 probes this relationship causally: can we \emph{induce} convergence by manipulating the task prompt, and do artifacts then restore exploration?

\subsection{Inducing Convergence via Prompt Pressure}

We apply four levels of convergence pressure to LRU cache (natural baseline = 0.540) and distributed scheduler (natural baseline = 0.310):

\begin{table}[h]
\centering
\caption{Baseline exploration under convergence pressure. $n = 20$ per cell.}
\label{tab:convergence_baselines}
\begin{tabular}{lcc}
\toprule
Pressure & LRU Cache & Dist.\ Scheduler \\
\midrule
None & 0.440 & 0.275 \\
Mild & 0.420 & 0.195 \\
Strong & 0.370 & 0.125 \\
Extreme & 0.440 & 0.105 \\
\bottomrule
\end{tabular}
\end{table}

Phase 3 ``none'' baselines (0.275 and 0.440) differ from the Phase 2 baselines in Table~\ref{tab:baselines} (0.310 and 0.540) because Phase 3 runs are fully independent replications. The differences (11\% and 19\%) are within the variance observed across 20-trial blocks throughout the study. All within-Phase-3 comparisons use the Phase 3 baselines.

\textbf{Distributed scheduler} shows a clean gradient: baseline drops from 0.275 to 0.105 as prompt pressure increases. Providing a code skeleton (extreme) cuts exploration by 62\% relative to the open-ended prompt. The manipulation works---we can induce convergence.

\textbf{LRU cache} shows a floor effect. Mild pressure drops baseline from 0.440 to 0.420, and strong pressure to 0.370, but extreme pressure does not push further (0.440). LRU cache has enough inherent design tension---the choice between data structures, thread safety strategies, and write policies---that agents cannot be fully suppressed below $\sim$0.37 coverage regardless of how prescriptive the prompt is.

\subsection{Artifacts Cannot Break Induced Convergence}

The critical test: do artifacts that help on naturally convergent tasks (rate limiter) also help on \emph{artificially} convergent tasks? We now have the full gradient---artifact effects at every pressure level for both tasks.

\textbf{Distributed scheduler---artifact effects across all pressure levels:}

\begin{table}[h]
\centering
\caption{Distributed scheduler artifact effects across convergence pressure levels. $n = 20$ per cell.}
\label{tab:dist_sched_gradient}
\small
\begin{tabular}{lccccccc}
\toprule
Pressure & Baseline & Anti-pat.\ $\Delta$ & Transcript $\Delta$ & Code $\Delta$ & Design $\Delta$ & Irrel.\ $\Delta$ & Avg $\Delta$ \\
\midrule
None & 0.275 & $-0.035$ & $-0.085$ & $-0.090$ & $-0.015$ & $-0.040$ & $-0.058$ \\
Mild & 0.195 & $+0.000$ & $-0.035$ & $-0.055$ & $+0.000$ & $+0.070$ & $+0.008$ \\
Strong & 0.125 & $+0.005$ & $+0.015$ & $-0.045$ & $-0.030$ & $+0.085$ & $+0.008$ \\
Extreme & 0.105 & $-0.025$ & $-0.055$ & $-0.025$ & $-0.015$ & $-0.010$ & $-0.027$ \\
\bottomrule
\end{tabular}
\end{table}

The gradient is unambiguous: as prompt pressure drives baseline from 0.275 down to 0.105, artifacts never become helpful. The average delta hovers near zero across all pressure levels. Compare this to the Phase 2 rate limiter---a naturally convergent task with baseline 0.033 where anti-patterns provide $+0.667$ and transcripts $+0.558$. The distributed scheduler at extreme pressure (baseline = 0.105) is within striking distance of rate limiter's baseline, yet anti-patterns provide $-0.025$ and transcripts $-0.055$. Same low baseline, opposite response.

\textbf{LRU cache---artifact effects across all pressure levels:}

\begin{table}[h]
\centering
\caption{LRU cache artifact effects across convergence pressure levels. $n = 20$ per cell.}
\label{tab:lru_cache_gradient}
\small
\begin{tabular}{lccccccc}
\toprule
Pressure & Baseline & Anti-pat.\ $\Delta$ & Transcript $\Delta$ & Code $\Delta$ & Design $\Delta$ & Irrel.\ $\Delta$ & Avg $\Delta$ \\
\midrule
None & 0.440 & $+0.060$ & $+0.030$ & $+0.090$ & $-0.010$ & $+0.090$ & $+0.042$ \\
Mild & 0.420 & $-0.030$ & $-0.060$ & $-0.130$ & $+0.090$ & $+0.100$ & $+0.000$ \\
Strong & 0.370 & $+0.010$ & $+0.030$ & $+0.000$ & $+0.060$ & $-0.010$ & $+0.018$ \\
Extreme & 0.440 & $-0.070$ & $-0.140$ & $-0.160$ & $-0.070$ & $-0.080$ & $-0.102$ \\
\bottomrule
\end{tabular}
\end{table}

LRU cache reveals a different dynamic. The baseline resists suppression---it never drops below 0.370 even under strong pressure, and rebounds to 0.440 under extreme pressure (the inherent design tension in LRU cache cannot be fully constrained). But the extreme prompt changes the \emph{response} to artifacts dramatically: from avg delta $+0.042$ (no pressure) to $-0.102$ (extreme pressure), even though both have baseline = 0.440. The prompt doesn't change how much agents explore; it changes how they respond to additional context. Every artifact hurts under extreme pressure, with code ($-0.160$) and transcript ($-0.140$) showing the strongest anchoring effects.

\textbf{The baseline-direction correlation breaks under manipulation.} In Phase 2, baseline exploration predicts artifact direction with $r = -0.821$ ($p < 0.001$) across 10 tasks with natural baselines. Across the 8 manipulated variants in Phase 3, the same correlation is $r = -0.067$---essentially zero. The predictor that works across natural tasks fails completely when convergence is artificially induced. This is the clearest evidence that natural and induced convergence are mechanistically different.

\subsection{Natural vs Induced Convergence}

The Phase 3 results reveal a mechanistic distinction that the Phase 2 correlation cannot capture. The collapse from $r = -0.821$ to $r = -0.067$ is not noise---it reflects two fundamentally different types of convergence that respond differently to artifacts.

\textbf{Natural convergence} arises from the model's training data. Rate limiter agents converge on token bucket because that is the dominant solution in the training corpus. This creates a \emph{soft prior}---the model defaults to it but can be dislodged. Anti-patterns work because they explicitly name alternatives (``don't just use token bucket''), which competes with the soft prior and forces deliberation. The prior is a habit, and habits can be disrupted.

\textbf{Induced convergence} arises from explicit instructions in the prompt. Distributed scheduler agents under extreme pressure converge because the prompt tells them to ``use this implementation.'' This creates a \emph{hard constraint}---the instruction is part of the task specification, not a default that can be overridden. Anti-patterns cannot dislodge an explicit instruction because the instruction outranks a suggestion about what not to do. The prompt is an order, not a habit.

The LRU cache results add a further wrinkle: induced convergence changes artifact response even when it does not change baseline exploration. At extreme pressure, LRU cache baseline remains 0.440 (identical to the no-pressure condition), but the average artifact effect swings from $+0.042$ to $-0.102$. The prompt pressure does not suppress exploration directly---the task's inherent design tensions resist that---but it changes \emph{how agents process additional context}. Under an extreme prescriptive prompt, any injected artifact compounds the constraint rather than disrupting it.

This three-way contrast---natural convergence (artifacts help), induced convergence with baseline suppression (artifacts neutral), induced convergence without baseline suppression (artifacts hurt)---suggests that the crossover effect in Phase 2 is a between-task phenomenon driven by training data priors, not a general property of any convergent system. Tasks naturally low in baseline exploration (rate limiter) converge because of soft priors that artifacts can disrupt. Tasks \emph{made} low in baseline exploration through prompt manipulation converge because of hard constraints that artifacts cannot override.

The practical implication is important: the baseline exploration diagnostic (Section~4.8) predicts artifact effectiveness specifically for \emph{natural} convergence---the kind that arises from model defaults, training data, and the inherent structure of the problem. It does not predict what happens when convergence is forced by system design (e.g., prescriptive prompts, constrained agent roles, or pre-specified architectures). Agent orchestration systems that use prescriptive role definitions or constrained prompts are inducing convergence---and should not expect knowledge artifacts to restore the exploration they suppressed.

\section{Discussion}

\subsection{Implications for Context Injection in Agent Systems}

The current default in agent orchestration is unconditional context injection: give agents all available relevant information. Our results show this is wrong for design tasks. The prescription is:

\begin{enumerate}
    \item \textbf{Measure baseline exploration first.} Run one or more no-context trials on the task and measure tradeoff coverage. This is a cheap diagnostic that predicts artifact direction ($r = -0.82$).

    \item \textbf{If baseline is low ($< 0.1$): inject disruption artifacts.} Anti-patterns and transcripts are most effective. Even irrelevant context helps. The goal is to break the default-solution anchor.

    \item \textbf{If baseline is moderate to high ($> 0.3$): do not inject relevant artifacts.} Relevant artifacts will narrow the design space more than they inform it. If any context is needed, prefer irrelevant or minimally-anchoring material.

    \item \textbf{Never assume relevant context helps.} On ML training pipeline, irrelevant context ($+0.087$) outperforms every relevant artifact. Relevance and helpfulness are not the same thing.
\end{enumerate}

\subsection{Implications for RAG and Long-Context Systems}

Retrieval-augmented generation retrieves \emph{relevant} documents and injects them into the prompt. Our results suggest this strategy is optimized for knowledge-recall tasks (where relevance ensures the right information is available) but may be counterproductive for design tasks (where relevance introduces harmful anchoring). RAG systems used in design contexts should consider \emph{anti-relevance}---deliberately surfacing alternative or contrarian perspectives rather than confirming the most likely approach.

\subsection{Why Design Is Different from Implementation}

Our results do not contradict the well-established finding that context improves code generation. Code generation is primarily a \emph{recall} task---the agent needs to know APIs, types, patterns, and conventions. Context provides this information. Design is primarily an \emph{exploration} task---the agent needs to consider alternatives and weigh tradeoffs. Context can constrain this exploration through anchoring.

The distinction is between convergent work (where you want the agent to find the right answer) and divergent work (where you want the agent to explore the space). Context helps convergent work and hurts divergent work. This mirrors the creativity literature on convergent vs divergent thinking \citep{guilford1967nature}.

\subsection{Limitations}

\begin{enumerate}
    \item \textbf{Single model family.} All experiments use Claude Sonnet 4. Convergence behavior and anchoring susceptibility may differ across model families. Cross-model validation is needed.

    \item \textbf{Tradeoff list subjectivity.} Known tradeoff lists were authored by the experimenters and validated against agent discussions. Different annotators might produce different lists, affecting absolute coverage numbers. Relative comparisons across conditions (within-task) are robust to this concern.

    \item \textbf{Agent-to-agent transfer only.} Human consumers may respond differently to the same artifacts. The anchoring mechanism may be weaker for humans with genuine domain expertise.

    \item \textbf{Design tasks only.} The crossover effect is demonstrated on software design problems. It likely does not apply to implementation tasks, debugging, or code review where context is informative rather than anchoring.

    \item \textbf{Self-referential evaluation.} Using Claude Sonnet 4 to evaluate Claude Sonnet 4's deliberation introduces a potential bias. Human annotation of a subset of trials would strengthen the evaluation methodology.

    \item \textbf{Temperature and team configuration.} All experiments use temperature 0.5 and 5-agent teams. Exploration patterns may differ at other configurations.

    \item \textbf{Exploration vs output quality.} We measure design space exploration (tradeoff coverage), not output quality (code correctness, architectural soundness). Higher exploration does not necessarily produce better systems---though it does produce more \emph{informed} decisions.
\end{enumerate}

\subsection{Future Work}

\begin{enumerate}
    \item \textbf{Cross-model validation.} Test with GPT-4, Gemini, and open-weight models to determine whether the crossover is a general LLM property or model-specific.

    \item \textbf{Human annotation study.} Hand-score a subset of trials to validate the LLM evaluator against human judgment.

    \item \textbf{Adaptive context injection.} Build an orchestrator that measures baseline exploration in real-time and decides whether to inject context. Measure whether this improves design quality over unconditional injection.

    \item \textbf{Anti-relevance retrieval.} Design a RAG variant that retrieves \emph{contrarian} or \emph{alternative} perspectives for design tasks instead of confirming the dominant approach.

    \item \textbf{Exploration-to-quality link.} Measure whether higher tradeoff coverage produces better downstream outcomes (fewer bugs, more maintainable code, better architectural fit).

    \item \textbf{Temporal dynamics.} Measure whether convergence is immediate (agents decide in the first few turns) or gradual (agents start exploring then collapse). This determines when disruption artifacts are most effective.
\end{enumerate}

\section{Conclusion}

We set out to study which knowledge artifacts best transfer design knowledge between agent teams. We found something more fundamental: context itself is not uniformly helpful.

On software design tasks, injected context acts as an anchor. When agents would otherwise converge on a default solution---because the model's training data contains a dominant answer---this anchoring disrupts the default and expands the design space. Anti-patterns increase rate limiter tradeoff coverage from 0.033 to 0.700, a 20$\times$ improvement ($p < 0.001$). Transcripts increase it by 17$\times$. Even an irrelevant document increases it by 10$\times$. Any context shakes agents loose from the default.

When agents would otherwise explore freely---because no dominant solution exists---this anchoring narrows the design space. Transcripts reduce Kubernetes operator coverage from 0.475 to 0.256, a 46\% degradation ($p < 0.001$). Code reduces it by 32\%. On ML training pipeline, an irrelevant document outperforms every relevant artifact. The content of the context matters less than whether agents were already exploring.

A single measurable variable---baseline exploration---predicts this direction with $r = -0.82$ ($p < 0.001$). The practical recommendation is not ``never inject context'' but ``measure before injecting.'' One no-context trial is a cheap diagnostic that tells you whether your knowledge artifacts will help or hurt.

We probed the mechanism further by manipulating convergence through prompt design. We can induce convergence---distributed scheduler baseline drops from 0.275 to 0.105 under extreme prompt pressure---but artifacts that rescue naturally convergent tasks fail to rescue artificially convergent ones. Natural convergence (training data priors) and induced convergence (explicit instructions) are mechanistically different. Artifacts can disrupt a model's default habits but cannot override explicit constraints.

The industry's assumption that more context is better is correct for implementation work. For design work---where the quality of exploration matters as much as the quality of output---context is a double-edged sword. Systems that treat context injection as a free lunch are leaving design quality on the table.

\bibliographystyle{plainnat}

\end{document}